\documentclass[conference]{IEEEtran}
\IEEEoverridecommandlockouts
\usepackage{cite}
\usepackage{amsmath,amssymb,amsfonts}
\usepackage{algorithmic}
\usepackage{algorithm}
\usepackage{graphicx}
\usepackage{textcomp}
\usepackage{xcolor}
\usepackage{multirow}
\usepackage{makecell}
\usepackage{booktabs}
\def\BibTeX{{\rm B\kern-.05em{\sc i\kern-.025em b}\kern-.08em
    T\kern-.1667em\lower.7ex\hbox{E}\kern-.125emX}}
\begin{document}

\title{CFMS: A Coarse-to-Fine Multimodal Synthesis Framework for Enhanced Tabular Reasoning}

\author{
    \textbf{Qixian Huang}\textsuperscript{$\ast$} \quad
    \textbf{Hongqiang Lin}\textsuperscript{$\ast$} \quad
    \textbf{Tong Fu} \quad
    \textbf{Yingsen Wang} \\
    \textbf{Zhenghui Fu} \quad
    \textbf{Qirui Wang} \quad
    \textbf{Yiding Sun}\textsuperscript{$\dagger$} \quad
    \textbf{Dongxu Zhang}\textsuperscript{$\dagger$} \\
    
}
\maketitle

\begin{abstract}
Reasoning over tabular data is a crucial capability for tasks like question answering and fact verification, as it requires models to comprehend both free-form questions and semi-structured tables. However, while methods like Chain-of-Thought (CoT) introduce reasoning chains, purely symbolic methodes are inherently limited by their blindness to holistic visual patterns. To address this, we propose the Coarse-to-Fine Multimodal Synthesis framework (CFMS), a novel two-stage paradigm that hierarchically decouples high-level visual perception from granular symbolic reasoning. In the Coarse Stage, CFMS leverages the Multimodal Large Language Models (MLLMs) to perform a one-time synthesis of a multi-perspective knowledge tuple. This tuple subsequently serves as a dynamic reasoning map to guide the fine stage, where a symbolic engine executes a targeted and efficient sequence of iterative operations over the table. Extensive experiments on the WikiTQ and TabFact benchmarks demonstrate that CFMS achieves competitive accuracy. The framework exhibits particular robustness when handling large tables and when instantiated with smaller backbone models, validating its effectiveness and generalizability.

\end{abstract}

{
  \renewcommand{\thefootnote}{\fnsymbol{footnote}}
  \footnotetext[1]{Equal contribution.}
  \footnotetext[2]{Corresponding authors.}
}


\section{Introduction}
\label{sec:intro}

Tables are a ubiquitous format for organizing and presenting structured information, playing a critical role in domains ranging from data analysis to scientific research\cite{zhang2025survey}. Consequently, the ability of artificial intelligence systems to reason over tabular data is of paramount importance\cite{ye2023large}. The advent of Large Language Models (LLMs) has demonstrated remarkable capabilities in natural language understanding\cite{deng2024tables}, yet their proficiency in interpreting the semi-structured nature of tables remains a significant challenge\cite{ye2023large,al2024findings}. These models often struggle to grasp the complex interplay between rows and columns that is essential for deep tabular reasoning.

\begin{figure}[t]
\centering
\includegraphics[width=0.85\linewidth]{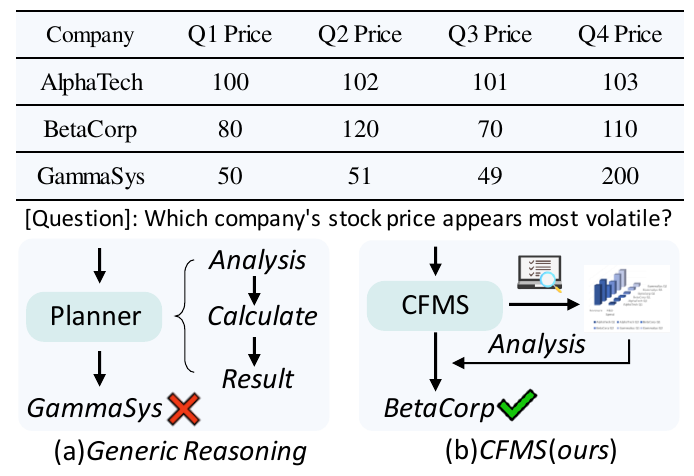}
\caption{An illustration of CFMS's advantage in holistic reasoning. (a) A symbolic-only method must resort to inefficient, brute-force calculations to identify volatility. (b) Our CFMS framework uses the Coarse Stage to synthesize a knowledge tuple $K_{coarse}$, including a visual perception $R_{visual}$ of a generated chart, which immediately identifies the holistic pattern and guides the Fine Stage to a correct and efficient answer.}
\label{fig1}
\end{figure}

To bridge this gap, recent advancements\cite{wang2024chain,jin2023tab} have explored several paradigms for symbolic reasoning. One direction is program-aided reasoning, exemplified by frameworks like Binder\cite{cheng2022binding}, which generate executable code to query a static table. Another direction involves modifying the table itself. Dater\cite{ye2023large} decomposes large tables using a fixed set of operations as a preprocessing step to fit within the model's context window. A more significant step forward is the Chain-of-Table\cite{wang2024chain} framework, which represents the reasoning process as a dynamic sequence of iterative table operations. However, this paradigm operates within a purely symbolic domain, treating tables as mere text.
This inherent symbolic blindness prevents the model from perceiving holistic patterns. For questions requiring visual understanding, such as identifying trends or outliers, models are forced to resort to inefficient brute-force computations.
When faced with a table of stock prices and the question ``Which company's stock price appears most volatile?" A symbolic-only model (Fig.~\ref{fig1}a) is blind to the visual pattern. It is forced to resort to inefficient calculations, such as computing the standard deviation or variance for every company in the table.

A natural evolution is to incorporate multimodal perception, leveraging the power of MLLMs to see the data in its entirety\cite{wang2503multimodal}. However, a naive end-to-end method, where large and complex table images are directly fed into a single MLLMs for iterative reasoning, presents substantial practical challenges. The significant computational overhead and GPU memory (VRAM) requirements for processing high-resolution visual data at each reasoning step can be prohibitive\cite{chen2025recent}, limiting the model's scalability and applicability to real-world and resource-constrained scenarios. This creates a critical dilemma: how can we harness the rich insights from multimodal perception without incurring the prohibitive costs of full-scale visual reasoning?

To address these challenges, we propose the CFMS framework, a hierarchical method that decouples high-level perception from detailed reasoning. Our motivation stems from the human cognitive process: first gaining a general overview, then delving into specifics. CFMS (Fig.~\ref{fig1}b) operationalizes this through a two-stage process.
In the initial stage, we leverage external MLLMs\cite{wang2503multimodal} to perform a high-level analysis of the table. This stage synthesizes a multi-perspective textual context, including descriptions of visual patterns, relevant semantic knowledge, and statistical summaries. By distilling multimodal insights into a textual format, this stage efficiently circumvents the memory bottleneck of continuous visual processing.
The second stage ingests the original table alongside the lightweight knowledge tuple generated in the coarse stage. This enhanced context guides a symbolic reasoning engine similar to Chain-of-Table\cite{wang2024chain} to execute an efficient and accurate sequence of table operations. The holistic understanding from the coarse stage directs the fine-grained reasoning process toward the most relevant parts of the data.

Our main contributions are summarized as follows:
\begin{itemize}
\item We propose the CFMS framework which decouples high-level multimodal perception from low-level symbolic reasoning to enrich the reasoning context.
\item We design an efficient knowledge synthesis mechanism that leverages MLLMs to perform a one-time distillation of tabular data.
\item We conduct extensive experiments on standard benchmarks, demonstrating that CFMS achieves competitive or superior performance.
\end{itemize}

\section{RELATED WORK}
\subsection{Multimodal Large Language Models }
MLLMs represent a significant advancement in artificial intelligence\cite{wang2503multimodal}, integrating the sophisticated reasoning capabilities of LLMs\cite{chen2025recent} with the perceptual acuity of vision models\cite{liu2023visual}. Therefore, MLLMs enable performing complex tasks that require a concurrent understanding of both images and text\cite{carbune2024chart}, such as visual question answering, image captioning and multimodal content generation\cite{zhang2025ascot,guo2025deepseek}.

The power of MLLMs lies in their ability to ground linguistic concepts in visual reality\cite{dai2023instructblip}, enabling them to reason about objects and their interactions\cite{zhang2023internlm} in a way that purely text-based models cannot. This progress has culminated in the development of powerful models like GPT-4V and a variety of open-source alternatives\cite{guo2025deepseek}. These models exhibit remarkable zero-shot\cite{wei2022chain} and few-shot capabilities\cite{chada2021fewshotqa}, demonstrating strong performance across a wide array of multimodal tasks\cite{wang2503multimodal,liu2023visual}.

\subsection{Chain-of-Thought Reasoning for Tabular Data}
The advent of CoT prompting marked a pivotal moment for LLMs\cite{zhang2025ascot,wei2022chain}, significantly enhancing their reasoning capabilities by enabling them to generate intermediate steps before arriving at a final answer. This core idea was extended by subsequent work, such as decomposing complex problems into simpler subproblems like Few-Shot QA and so on\cite{chada2021fewshotqa,zhang2026chain,zhang2026not}. However, these generic reasoning frameworks are not specifically designed for the structured nature of tabular data\cite{zhang2025survey}, often failing to effectively leverage the relational information inherent in rows and columns\cite{ni2023lever}.

To bridge this gap, research has evolved in two primary directions\cite{wang2024chain}. One prominent line of work involves program-aided reasoning\cite{ye2023large}, where LLMs are prompted to generate executable code, such as Python or SQL to query the table\cite{cheng2022binding}. This method excels at tasks requiring precise numerical calculations and has been shown to improve arithmetic reasoning significantly.
A more advanced paradigm is iterative table transformation\cite{gao2023pal}. This method fundamentally extends the CoT concept by using the table itself as the medium for intermediate reasoning\cite{wei2022chain}. Instead of generating textual thoughts, the LLM is guided to produce a sequence of programmatic table operations. Each operation is then executed, creating a new and transformed table that is fed back into the model as the context for the next reasoning step.

Concurrently, research in the multimodal domain has focused on enabling MLLMs to understand table images directly\cite{liu2023visual}. Works like LLaVA\cite{lin2023video} have shown that domain-specific instruction tuning is essential for teaching MLLMs to parse visual table structures\cite{deng2024tables,wang2024chain}. 

\begin{figure*}[t]
\centering
\includegraphics[width=0.95\linewidth]{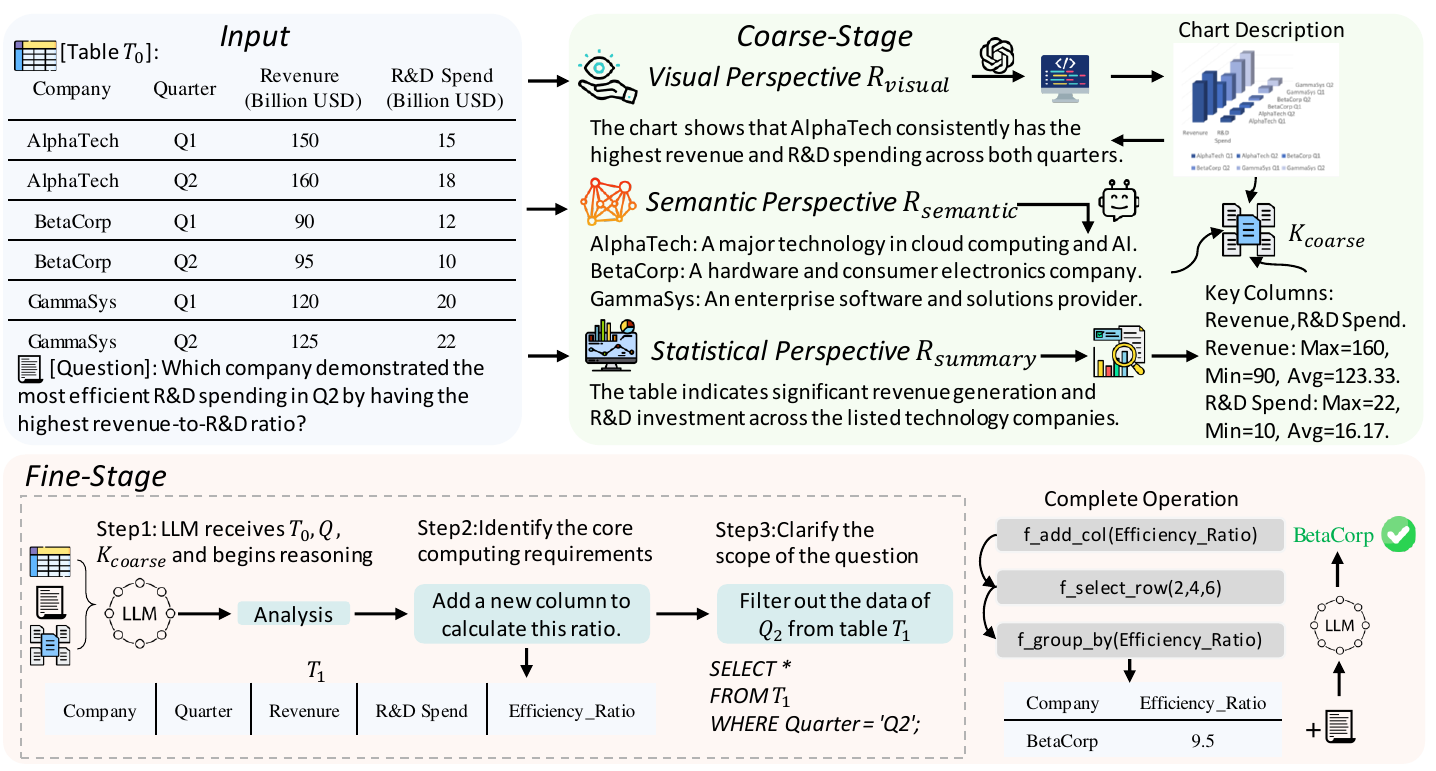}
\caption{An overview of the CFMS framework. The process begins with the Coarse Stage, where the initial table $T_0$ is analyzed from multiple perspectives (visual, semantic, statistical) to synthesize a textual knowledge tuple $K_{coarse}$. Subsequently, in the Fine Stage, an LLM planner leverages this holistic context to guide a step-by-step iterative reasoning process, executing a sequence of precise table operations to derive the final answer.}
\label{fig2}
\end{figure*}

\section{METHODOLOGY}

\subsection{Problem Formulation}

\begin{algorithm}[tb]
\caption{Coarse-to-Fine Multimodal Synthesis (CFMS)}
\label{alg}
\textbf{Input}: Initial table $T_0$, Question $Q$\\
\textbf{Output}: Predicted answer $\hat{A}$
\begin{algorithmic}[1]
\STATE $K_{\mathrm{coarse}} \gets \mathrm{GenerateKnowledge}(T_0)$ 
\STATE $T \gets T_0$
\WHILE{not termination condition}
    \STATE $(o, args) \gets \mathrm{PlanOperation}(T, Q, K_{\mathrm{coarse}})$
    \IF{$o = \mathrm{<END>}$}
        \STATE \textbf{break}
    \ENDIF
    \STATE $T \gets \mathrm{Execute}(o, args, T)$
\ENDWHILE
\STATE $\hat{A} \gets \mathrm{GenerateAnswer}(T, Q, K_{\mathrm{coarse}})$
\STATE \textbf{return} $\hat{A}$
\end{algorithmic}
\end{algorithm}

Given an initial table $T_0$ and an associated question $Q$, our objective is to learn a function $f$ that maps the input pair ($T_0$, $Q$) to the answer $A$. The CFMS framework operationalizes this function through a hierarchical, two-stage process. It first transforms the singular ($T_0$, $Q$) input into a rich, multi-perspective knowledge context, upon which subsequent fine-grained reasoning is performed. The overall workflow is illustrated in Fig.~\ref{fig2}. We also provide Algorithm~\ref{alg} for CFMS.

\subsection{Coarse-Grained Knowledge Enhancement}

The objective of this stage is to construct a multi-dimensional textual context. We leverage external generative models to produce a set of knowledge-enhanced descriptions from three complementary perspectives in parallel.

For the visual modality, we employ a two-step procedure to textualize the visual information within the table.
First, the table $T_0$ is combined with a prompt $P_{\text{chart}}$ and input into a code-generating LLM $\mathcal{G}_{\text{code}}$ to generate visualization code. The model selects the most suitable chart format and the execution of this code yields a chart $C_0$
~\cite{zhang2026igasa,sun2026align}. Subsequently, the resulting chart $C_0$ is fed into a MLLMs $\mathcal{M}_{\text{vlm}}$ along with a descriptive prompt $P_{\text{visual\_desc}}$ to generate a textual description $R_{\text{visual}}$, which captures the key visual patterns:
\begin{equation}
    R_{\text{visual}} = \mathcal{M}_{\text{vlm}}(C_0 \oplus P_{\text{visual\_desc}}),
\end{equation}
where $\oplus$ denotes the concatenation of input content.

For the semantic modality, we aim to ground the symbolic entities within the table in real-world context. To this end, we generate a semantic rationale $R_{\text{semantic}}$ by employing a knowledge generation model $\mathcal{G}_{\text{kg}}$, in conjunction with a prompt $P_{\text{semantic}}$. This process extracts commonsense knowledge related to key entities present in $T_0$:
\begin{equation}
    R_{\text{semantic}} = \mathcal{G}_{\text{kg}}(T_0 \oplus P_{\text{semantic}}).
\end{equation}

For the statistical modality, we generate a textual summary $R_{\text{summary}}$ to equip the model with a quantitative summary of the table's contents. This is achieved by employing a standard LLM $\mathcal{G}_{\text{llm}}$ guided by a CoT prompt $P_{\text{summary}}$. The prompt directs the model to first identify key numerical columns, calculate their essential statistics ($e.g.$, maximum, minimum, average) and then synthesize these findings into a concise natural language conclusion:
\begin{equation}
    R_{\text{summary}} = \mathcal{G}_{\text{llm}}(T_0 \oplus P_{\text{summary}}).
\end{equation}

Collectively, the descriptions generated from these complementary perspectives are aggregated to form a final knowledge tuple $K_{\text{coarse}}$. This tuple encapsulates a multimodal understanding of the initial table in a compact textual format, guiding the LLM planner to execute more targeted operations for the fine-grained reasoning stage that follows.

\subsection{Fine-Grained Iterative Reasoning}

This stage functions as the core reasoning engine of the framework, initiating a knowledge-guided and iterative table evolution process that extends the core mechanism of Chain-of-Table. It takes the original table $T_0$ and the knowledge tuple $K_{\text{coarse}}$ as input, transforming the static table into a dynamic reasoning canvas~\cite{wang2026personalq,zhang2026cmhanet}.

At any timestep $t$, the reasoning state is defined by the current table $T_t$ and the history of operations $H_t$, denoted as $S_t = (T_t, H_t)$. The central mechanism is a knowledge-guided dynamic planning step, where an LLM-based planning policy $\pi_{\text{llm}}$, selects the subsequent table operation $o_{t+1}$ and its corresponding arguments $args_{t+1}$. This decision is based on the current state $S_t$, the question $Q$, and the tuple $K_{\text{coarse}}$:
\begin{equation}
    (o_{t+1}, args_{t+1}) \sim \pi_{\text{llm}}(S_t, Q, K_{\text{coarse}}).
\end{equation}

Here, $o_{t+1}$ denotes the selected atomic operation and $args_{t+1}$ represents the specific parameters required to execute it. Following this step, the selected operation is applied to the current table $T_t$ to produce the subsequent table state $T_{t+1}$:
\begin{equation}
    T_{t+1} = o_{t+1}(T_t, args_{t+1}).
\end{equation}

The operation history is updated accordingly: $H_{t+1} = H_t \oplus (o_{t+1}, args_{t+1})$. 
The fine-grained planner $\pi_{llm}$ selects operations from a predefined symbolic pool. This set is informed by previous work~\cite{wang2024chain}. It includes essential operations for table manipulation such as $f\_add\_column$ to create new computed columns, $f\_select\_column$ to subset columns, $f\_select\_row$ to filter rows, and $f\_group\_by$, $f\_sort\_by$ for aggregation or ordering.
This iterative process continues until the planner generates a termination token  $\mathrm{<END>}$.
At each step $t$, the planner model receives a consolidated prompt containing the general question $Q$, the complete knowledge tuple $K_{coarse}$, and the current serialized table $T_t$. This guides the planner to select the most relevant operation $o_{t+1}$ from the symbolic pool that aligns with the high-level insights, rather than resorting to brute-force exploration or sampling.


\subsection{Final Answer Generation}




Upon the termination of the fine-grained reasoning loop, the process produces a final evolved table $T_{final}$. This table encapsulates the precise data required to address the query. To generate the definitive answer, this final table $T_{final}$ is concatenated with the original question $Q$ and the complete knowledge tuple $K_{coarse}$. This comprehensive final context $C_{final}$ is then inputted into the language model $\mathcal{G}_{llm}$ to produce the final answer $A$:
\begin{equation} 
A = \mathcal{G}_{\text{llm}}(C_{\text{final}}). 
\end{equation} 

The model synthesizes all available information including the original query, the coarse-grained holistic knowledge, and the results of the fine-grained symbolic operations to formulate its definitive response.

\section{EXPERIMENTS}
\subsection{Experimental Settings}
\textbf{Datasets}\quad We evaluated the effectiveness of our CFMS framework on two established public benchmarks: WikiTQ\cite{pasupat2015compositional} and TabFact\cite{chen2019tabfact}.
WikiTQ is a question-answering dataset that necessitates complex reasoning over tables to derive a short text span as the answer. Following the official guidelines, we report performance using denotation accuracy.
TabFact serves as a benchmark for table-based fact verification. The objective is to determine whether a given textual statement is true or false based on the provided table. For this task, we measure performance using binary classification accuracy.

\textbf{Implementation Details}\quad For the coarse stage, we utilize GPT-4V to generate visual descriptions from charts and GPT-4\cite{achiam2023gpt} to produce semantic and statistical summaries.
For the fine stage, we employ a suite of powerful LLMs to serve as the backbone planner. Our experiments utilize GPT-3.5-Turbo and LLaMA-2-13B-Chat to ensure a complete evaluation across different model architectures.
All experiments are conducted using a few-shot and in-context learning method.

\textbf{Baseline Models}\quad We compare our CFMS framework against two main categories. 
Generic Reasoning category includes methods that treat the table as serialized text and rely on the LLM's general reasoning abilities. Baselines in this group include End-to-End QA, Few-Shot QA\cite{brown2020language}, and Chain-of-Thought\cite{wei2022chain}.
Program-Aided Reasoning leverages more structured methods, either by generating code or directly manipulating the table. Baselines in this group include Text-to-SQL\cite{rajkumar2022evaluating}, Binder\cite{cheng2022binding}, and Dater\cite{ye2023large}.

\begin{table}[t]
\centering
\caption{Table understanding results on TabFact and WikiTQ datasets across GPT 3.5 and LLaMA 2 models. We report accuracy (\%).}
\begin{tabular}{l 
                *{2}{c} 
                *{2}{c}}
\toprule
\multirow{2}{*}{\textbf{Methods}} 
  & \multicolumn{2}{c}{\textbf{GPT 3.5}} 
  & \multicolumn{2}{c}{\textbf{LLaMA 2}} \\
\cmidrule(lr){2-3} \cmidrule(lr){4-5}
  & \textbf{TabFact} & \textbf{WikiTQ} 
  & \textbf{TabFact} & \textbf{WikiTQ} \\
\midrule
End-to-End QA & 70.45 & 51.84 & 44.86 & 23.90 \\
Few-Shot QA & 71.54 & 52.56 & 62.01 & 35.52 \\
Chain-of-Thought & 65.37 & 53.48 & 60.52 & 36.05 \\
Chain-of-Table & 80.20 & 59.94 & 67.24 & 42.61 \\
Text-to-SQL & 64.71 & 52.90 & 64.03 & 36.14 \\
Binder & 79.17 & 56.74 & 62.76 & 30.92 \\
Dater & 78.01 & 52.81 & 65.12 & 41.44 \\
\midrule
CFMS (ours) & 
\textbf{81.37} &
\textbf{61.42} &
\textbf{68.53} &
\textbf{43.89} \\
\bottomrule
\end{tabular}
\label{tab1}
\end{table}

\begin{table}[t]
\centering
\small
\caption{Performance of Related Methods on small, medium, and large tables from WikiTQ. We also report accuracy (\%).}
\renewcommand{\arraystretch}{0.9} 
\resizebox{0.8\linewidth}{!}{
\begin{tabular}{lccc}
\toprule
\multirow{2}{*}{\textbf{Methods}}  & 
\multicolumn{3}{c}{\textbf{Table Size}} \\
\cmidrule(lr){2-4}
 & \textbf{Small} & \textbf{Medium} & \textbf{Large} \\
\midrule
Binder & 56.54 & 26.13 & 6.41 \\
Dater & 62.50 & 42.34 & 34.62 \\
CFMS (ours) & 
\textbf{69.26} & 
\textbf{53.47} & 
\textbf{46.92} \\
\bottomrule
\end{tabular}
}
\label{tab2}
\end{table}

\subsection{Overall Results}

We evaluated our proposed CFMS framework on WikiTQ and TabFact. The main results are presented in Table~\ref{tab1}.
Our method outperforms all baseline methods in both datasets and both backbone models.
From the results, we observe that CFMS consistently delivers state-of-the-art performance. For instance, our framework achieves an accuracy of 81.37\% on TabFact and 61.42\% on WikiTQ with GPT 3.5. Compared to an End-to-End QA method, CFMS yields a performance increase of 10.92\% on TabFact and 9.58\% on WikiTQ.

Furthermore, we observe that the architectural benefits of CFMS are evident across LLMs of varying scales. While the performance of all methods decreases substantially when shifting from GPT 3.5 to the smaller LLaMA 2 model, CFMS maintains its SOTA position relative to the baselines. On WikiTQ, CFMS (68.53\%) still outperforms Dater (65.12\%) and Binder (62.76\%), which suggests the structural advantage of the CFMS framework is robust.

\begin{table}[t]
\centering
\caption{Number of queries generated for a single question in the proposed CFMS on the WikiTQ dataset.}
\resizebox{0.90\columnwidth}{!}{%
\begin{tabular}{l c c}
\toprule
\textbf{Methods} & \textbf{\makecell{Total \# of \\ queries}} & \textbf{\makecell{\# of queries\\ in each step}} \\
\midrule
Binder & 50 & Generate Program: 50 \\
\midrule
Dater & 100 & \makecell{Decompose Table: 40;\\ Generate Cloze: 20; \\ Generate SQL: 20; Query: 20} \\
\midrule
CFMS (ours) & $\leq$15 & \makecell{Coarse Stage: 4; \\Fine Stage: $\leq$10; \\ Final Query: 1} \\
\bottomrule
\end{tabular}}
\label{tab3}
\end{table}

To assess the robustness of our framework on tables of various sizes, we categorize the test samples from WikiTQ into three groups based on their token count: small ($<$2000 tokens), medium (2000 to 4000 tokens), and large ($>$4000 tokens). We compare CFMS with Dater and Binder. The detailed results are presented in Table~\ref{tab2}. The performance of all methods decreases as the input table size increases. Nevertheless, CFMS demonstrates superior resilience. While Dater's accuracy drops by 27.9 points on large tables, CFMS's accuracy drops by only 22.3 points. This robustness in complex and long-context scenarios increases CFMS's absolute accuracy advantage over Dater from 6.76\% on small tables to 12.3\% on large tables, validating the effectiveness of our coarse-to-fine method.

\begin{figure}[t]
\centering
\includegraphics[width=0.85\linewidth]{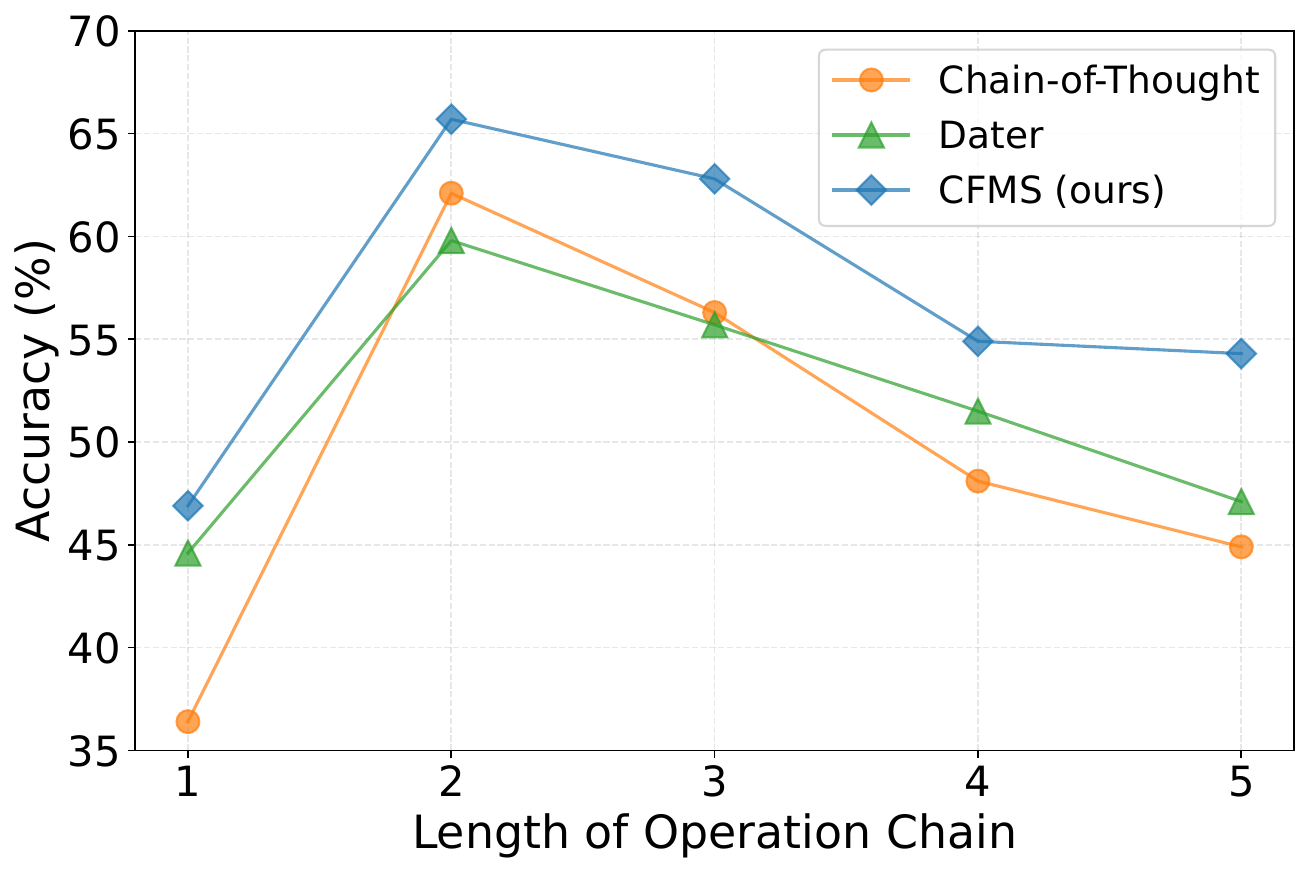}
\caption{Performance of the proposed CFMS on WikiTQ for questions that require an operation chain of varying lengths.}
\label{fig3}
\end{figure}

In the fine-grained reasoning stage of our CFMS framework, the number of operations is not fixed but is dynamically determined based on the complexity of the question and the table. Therefore, we conduct a detailed study on the performance under different numbers of operations by categorizing the test samples from WikiTQ. For each chain length, we plot the accuracy of all methods in Fig.~\ref{fig3}. Notably, CFMS consistently surpasses both baseline methods across all operation chain lengths, maintaining a significant performance margin.

\subsection{Analysis}
\textbf{Case Study}\quad
In Fig.~\ref{fig2}, we illustrate the end-to-end reasoning process of our CFMS framework on a representative task. The question poses a complex analytical challenge, which requires the model to first interpret the abstract concept of ``R\&D efficiency" as a ``revenue-to-R\&D ratio." Subsequently, it must calculate this new metric for each entry, filter the results to the relevant quarter ($Q2$), and ultimately identify the company with the maximum ratio.
Our proposed CFMS framework first addresses this in the Coarse Stage. Instead of immediately performing symbolic operations, it synthesizes a multi-perspective knowledge tuple ($K_{coarse}$). This initial step provides the model with a crucial high-level understanding.

This holistic context then guides the Fine Stage. The LLM planner does not perform a blind search. Its first action is to address the central requirement of the question by planning an $add\_column$ operation to compute the ``Efficiency\_Ratio". Subsequently, it narrows the scope by executing a $select\_row$ operation to filter for the ``$Q2$" data before finding the final answer. The intermediate tables, therefore, represent a thought process that has been intelligently directed from the outset.

\textbf{Efficiency Analysis}\quad 
We analyze the efficiency of our CFMS framework by evaluating the number of LLM queries required to answer a single question, as presented in Table~\ref{tab3}. This metric reflects the complexity and efficiency of the reasoning path rather than the raw computational cost of individual queries. Methods like Binder and Dater often rely on large-scale self-consistency sampling to boost performance, which involves generating a large number of candidate outputs to ensure robustness.
While effective, this brute-force sampling method is inefficient.
In contrast, CFMS introduces a small number of queries in the Coarse Stage to synthesize the $K_{coarse}$. This initial synthesis provides high-level guidance that makes the subsequent Fine Stage remarkably targeted. The guided Fine Stage requires $\le10$ deterministic operations, bringing the total query count to $\le15$.
We acknowledge that the Coarse Stage queries represent a higher cost and latency per query than the simpler baseline queries. However, this is a one-time investment. This trade-off is favorable, as it allows CFMS to achieve superior accuracy and robustness while drastically reducing the total number of reasoning steps.

\begin{table}[t]
\centering
\small
\caption{Ablation study of the CFMS framework on TabFact and WikiTQ datasets. We report Accuracy (\%).}
\renewcommand{\arraystretch}{0.9} 
\resizebox{0.8\linewidth}{!}{ 
\begin{tabular}{lcc}
\toprule
\multirow{2}{*}{\textbf{Methods}} & 
\multicolumn{1}{c}{\textbf{TabFact}} & 
\multicolumn{1}{c}{\textbf{WikiTQ}} \\
\cmidrule(lr){2-3}
 & \textbf{Accuracy} & \textbf{Accuracy} \\
\midrule
\textsc{CFMS} & \textbf{81.37} & \textbf{61.42} \\
w/o $f\_add\_column()$ & 80.10$_{(1.27\downarrow)}$ & 58.86$_{(2.56\downarrow)}$ \\
w/o $f\_select\_column()$ & 77.61$_{(3.76\downarrow)}$ & 58.29$_{(3.13\downarrow)}$ \\
w/o $f\_select\_row()$ & 76.79$_{(4.58\downarrow)}$ & 57.65$_{(3.77\downarrow)}$ \\
w/o $f\_group\_by()$ & 79.43$_{(1.94\downarrow)}$ & 55.02$_{(6.40\downarrow)}$ \\
w/o $f\_sort\_by()$ & 80.88$_{(0.49\downarrow)}$ & 58.61$_{(2.81\downarrow)}$ \\
\bottomrule
\end{tabular}
}
\label{tab4}
\end{table}

\begin{table}[t]
\centering
\small
\caption{Ablation study of the core components of our CFMS framework on TabFact and WikiTQ datasets.}
\renewcommand{\arraystretch}{0.9}
\resizebox{0.8\linewidth}{!}{
\begin{tabular}{lcc}
\toprule
\multirow{2}{*}{\textbf{Methods}} & 
\multicolumn{1}{c}{\textbf{TabFact}} & 
\multicolumn{1}{c}{\textbf{WikiTQ}} \\
\cmidrule(lr){2-3}
 & \textbf{Accuracy} & \textbf{Accuracy} \\
\midrule
CFMS (Full Model) & \textbf{81.37} & \textbf{61.42} \\
\midrule
w/o Coarse Stage & 73.55$_{(7.82\downarrow)}$ & 55.18$_{(6.24\downarrow)}$ \\
w/o Fine Stage & 75.96$_{(5.41\downarrow)}$ & 56.44$_{(4.98\downarrow)}$ \\
\bottomrule
\end{tabular}
}
\label{tab5}
\end{table}

\subsection{Ablation Study}

To demonstrate the effectiveness of our method, we performed several ablation studies. We remove one of the pre-defined operations ($e.g.$, $f\_add\_column()$, $f\_select\_row()$, etc.) from the available operation pool. Consequently, the LLM planner must construct its reasoning chain using only the remaining operations. We report the results of this ablation study in Table~\ref{tab4}.
As we can see, all operations contribute to the final state-of-the-art performance, as removing any single operation results in a notable decrease in accuracy on both datasets. We also observe that $f\_select\_row()$ and $f\_select\_column()$ are the most critical operations for the TabFact benchmark, with performance drops of 4.58\% and 3.76\%, respectively.

To validate the contributions of the two core components of our CFMS framework. We conducted a comprehensive ablation study. We report the results of this study in Table~\ref{tab5}. As shown, the removal of either stage leads to a significant degradation in performance, confirming that both are integral to the success of our framework. The substantial drop in accuracy for the w/o Coarse Stage variant underscores the importance of the initial knowledge synthesis. This study also validates our core motivation: the synergistic combination of high-level perception (Coarse Stage) and detailed manipulation (Fine Stage) is key to achieving state-of-the-art performance in complex tabular reasoning.

\section{CONCLUSION}

In this work, we propose the CFMS framework, a novel hierarchical system that effectively decouples high-level perception from detailed symbolic reasoning. CFMS first employs a Coarse Stage, where powerful MLLMs perform a one-time synthesis from visual, semantic, and statistical viewpoints. This textual ``reasoning map" then guides a Fine Stage of iterative, symbolic table transformations. Our extensive experiments on the WikiTQ and TabFact benchmarks demonstrate that CFMS achieves strong performance. While acknowledging the computational trade-off in the Coarse Stage and the potential for error propagation in the reasoning map, these areas highlight critical directions for future work in enhancing efficiency and robustness. This research advocates for a shift from monolithic reasoning paradigms to hierarchical, coarse-to-fine strategies that mimic human cognition. We hope our investigations pave the way for future advancements~\cite{wang2023enhancing,li2023synergy,xue2024integrating} in multimodal integration and enhance the reliability, robustness, and effectiveness of LLM reasoning on structured data~\cite{zhang2026pointcot,wang2026pointrft,sun2025hyperpoint}.

\bibliographystyle{IEEEbib}
\bibliography{icme2025references}

\vspace{12pt}

\end{document}